\pdfoutput=1

\documentclass[11pt]{article}

\usepackage[]{naacl2021}

\usepackage{times}
\usepackage{latexsym}
\usepackage{amsmath}
\usepackage{graphicx}
\usepackage{stfloats}
\usepackage{multirow}
\usepackage{amssymb}

\usepackage[noend]{algpseudocode}
\usepackage{algorithmicx,algorithm}
\usepackage{verbatim}

\usepackage[T1]{fontenc}

\usepackage[utf8]{inputenc}

\usepackage{microtype}

\title{RTFE: A Recursive Temporal Fact Embedding Framework for \\Temporal Knowledge Graph Completion
}

\author{Youri Xu, Haihong E\thanks{$^*$Corresponding author: Haihong E} , Meina Song, wenyu song, Xiaodong Lv, wang haotian, yang jinrui\\
  School of Computer Science, Beijing University of Posts and Telecommunications,  China \\
  \texttt{\{youri.xu,ehaihong,mnsong,swy9834,lvxiaodong,haotianwang,yangjinrui\}}\\ \texttt{@bupt.edu.cn} \\}

\begin{document}
\maketitle
\begin{abstract}
Static knowledge graph (SKG) embedding (SKGE) has been studied intensively in the past years. Recently, temporal knowledge graph (TKG) embedding (TKGE) has emerged. In this paper, we propose a Recursive Temporal Fact Embedding (RTFE) framework to transplant SKGE models to TKGs and to enhance the performance of existing TKGE models for TKG completion. Different from previous work which ignores the continuity of states of TKG in time evolution, we treat the sequence of graphs as a Markov chain, which transitions from the previous state to the next state. RTFE takes the SKGE to initialize the embeddings of TKG. Then it recursively tracks the state transition of TKG by passing updated parameters/features between timestamps. Specifically, at each timestamp, we approximate the state transition as the gradient update process. Since RTFE learns each timestamp recursively, it can naturally transit to future timestamps. Experiments on five TKG datasets show the effectiveness of RTFE.
\end{abstract}

\section{Introduction}

\begin{figure*}[t]
    \centering
    \includegraphics[width=1\textwidth]{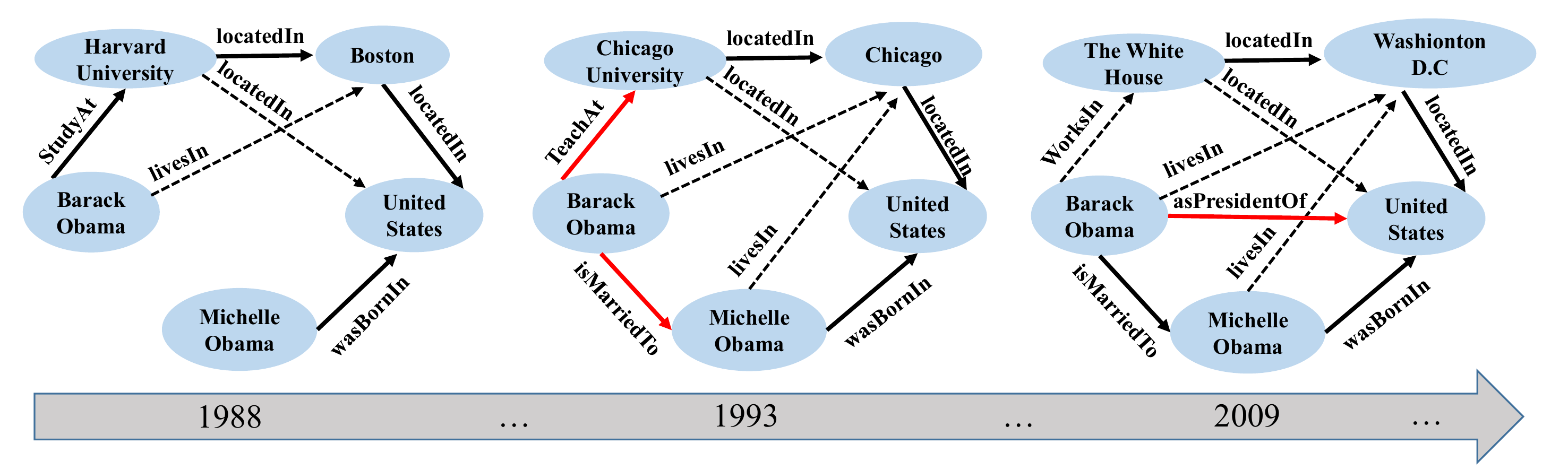}
    \caption{A toy example of TKG where solid edges represent observed edges and red edges represent new facts that occurred at that timestamp. Besides, dotted edges represent missing or potential facts.}
    \label{fig:TKG+problem_defi}
\end{figure*}
Temporal knowledge graph (TKG) is an extension of static knowledge graphs (SKGs) which introduce the time dimension. In SKGs, facts are considered to be time-invariant (\citealp{sil2014towards}). In reality, facts are not always true. For example, the triple \textit{(Obama, President, United States)} was true only from 2009 to 2016 and \textit{(Obama, married, Mitchell)} since 1992. However, SKGs do not reflect the change in facts over time. An example of TKG is shown in Figure \ref{fig:TKG+problem_defi}. Besides, facts on social networks, e-commerce platforms and trading platforms also change over time. Therefore, TKGs have the potential to improve the performance of question answering, search, recommendation and prediction based on KGs (\citealp{huang2020cluster}; \citealp{garg2020temporal}).

TKG can be expressed as a set of quadruples \textit{(subject, relation, object, timestamp)}. Different from SKGs which ignore the time attribute of facts, the facts of TKGs are distributed in timestamps, which can reflect the dynamic change of entities and relationships over time. Due to the limited coverage of KGs, TKGs are also incomplete. By completing TKG, missing and potential knowledge under specific timestamps can be found.

In recent years, a lot of work (\citealp{bordes2013translating}; \citealp{wang2014knowledge}; \citealp{linlearning}; \citealp{kazemi2018simple}; \citealp{schlichtkrull2018modeling}; \citealp{sun2019rotate};  \citealp{zhang2020learning}) has focused on KG completion by methods of graph embedding. These efforts have yielded good results, but most of them focused on SKGs and required training in a large number of triples.

However, the TKG under a certain timestamp is a sparse multi-relation graph (\citealp{esteban2016predicting}), so it is necessary to absorb information from other timestamps. What’s more, SKGE methods lacked the modeling of time attribute of relations, and were proposed based on the assumption that all facts occur at the same time. So they cannot reflect the temporal dependencies of facts. To handle these two problems, our RTFE passes parameters and features between timestamps in a recursive manner, which not only alleviates the sparsity problem of TKG, but takes advantage of the continuity and relevance characteristics of the fact as well.

Existing TKG completion methods (\citealp{dasgupta2018hyte};  \citealp{goel2019diachronic}; \citealp{lacroix2020tensor}) follow the training pattern of SKGE, which shuffles facts \textit{(s, r ,o ,t)} of different time randomly and learns all facts in a chaotic temporal order by mini-batch gradient descent algorithm. In other words, they just take time as a parameter but ignore the correlations in time evolution.

However, the early state may affect the later one and later facts tend to be dependent on early ones. In particular, the state at $t_i$ directly influences that at $t_{i+1}$. E.g., \textit{(Obama, Campaign, President, 2008)} directly influences \textit{(Obama, Inaugurated as, President, 2009)}. It has been verified that the chronological order of events can be used to improve the performance of link prediction (\citealp{jiang2016towards}; \citealp{jiang2016encoding}). Based on this, we further find that the early training state can improve later one if we train facts in their chronological order. In order to capture changes in TKG’s state transition, we think TKG as a sequence of dynamic graphs, not as a whole graph labeled with time information.

Besides, since new facts of future timestamps can be added to TKG, TKG is expanding dynamically. And the graphs of new timestamps may still be incomplete. However, existing TKG completion methods provide no solution to complete unseen future graphs. In their training pattern, facts of all timestamps are trained jointly to complete the graphs that have appeared. Models may need to be retrained on facts of all timestamps when a new timestamp appears. In contrast, our RTFE embeds and completes TKG’s each timestamp in a recursive way. By using the information of previous timestamps, RTFE can be naturally extended to future timestamps during the state transition of parameters/features. RTFE only needs to be trained on new emerging facts, which is light and immediate.

SKGE has been studied for many years while TKGE is still at birth. Problems encountered in SKGE (e.g., diverse relation patterns) can also occur in TKGE. Thus the advantages of SKG researches can be used to accelerate the development of TKGs if we bridge the gap between them. Our RTFE provides a way to migrate SKGE methods to TKGs while preserving their excellent effects.

Further, existing TKG completion methods designed specifically for the characteristics of TKGs can also be enhanced using the training pattern of RTFE. To sum up, we have made the following contributions:
\begin{enumerate}
    \item We propose a training pattern to bridge the gap between SKGE and TKGE. Therefore, state-of-the-art SKGE models can be used to accelerate the development of TKGE.
    \item Existing TKGE models can be further enhanced with our framework RTFE, after finishing their own regular training.
    \item To the best of our knowledge, we are the first to deal with the TKG evolution problem (i.e., new future timestamps are added to TKGs) in the TKG completion task.
    \item The experimental results on 5 TKG datasets show that RTFE preserve SKGE models’ excellent performance. And the predictive performance of state-of-the-art TKGE models are further enhanced using RTFE.
\end{enumerate}

\section{Problem definition}

\begin{figure*}[t]
    \centering
    \includegraphics[width=1\textwidth]{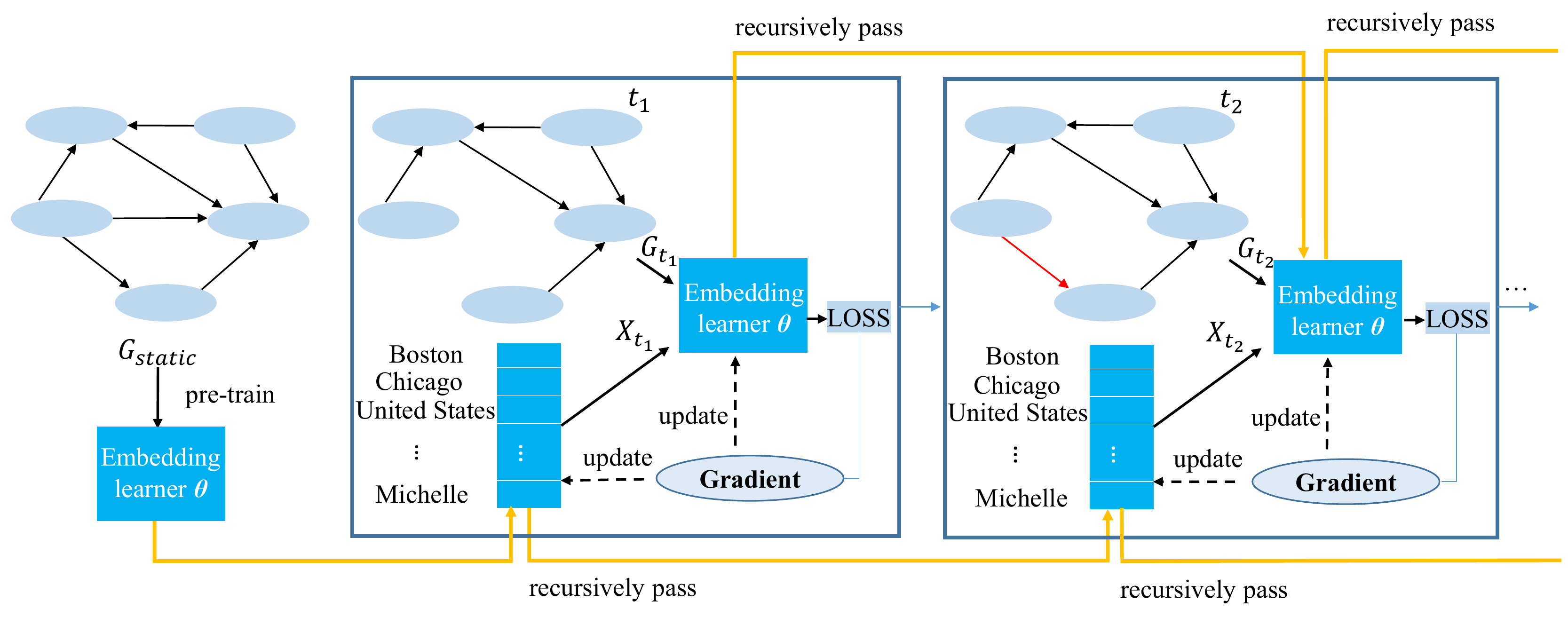}
    \caption{The framework of RTFE. TKG is first transformed to SKG $G_{static}$. RTFE pre-train embedding learner $\theta$ on $G_{static}$ to obtained the input of the first timestamp. Then features and parameters are recursively passed to next timestamp after learning the current timestamp.}
    \label{fig:RTFE}
\end{figure*}
A temporal knowledge graph (TKG) can be represented as a sequence of graphs, i.e. $G=\{G_{t_1},…,G_{t_n}\}$ where $G_{t_i}$ is a set of quadruples that occured at timestamp $t_i$, i.e. $G_{t_i}=\{(s,r,o,t_i )\}$ where $V$ is the set of $G$’s entities and $s,o \in V$; $R$ is the set of $G$’s relations and $r \in R$.

We focus on the following task: given a training TKG $G_{train}=\{G_{t_1},…,G_{t_n}\}$ , to infer the missing quadruples $(s, r, o, t)$ in test set $G_{test}=\{G_{t_1}^{'},…,G_{t_n}^{'}\}$ (i.e., assign high scores to true quadruples and low scores to false ones). As shown in Figure \ref{fig:TKG+problem_defi}, missing facts with high probability are dotted.

\section{Recursive Temporal Fact Embedding (RTFE) Framework}

The state of TKGs change with the change of entities and relations over time. SKGE models fail to capture correlations during state transition. And existing TKGE models for TKG completion capture it implicitly. It can be observed that the TKG after the current change changes on the closest former state, which is similar to a first order Markov chain (Given the state at the current moment, the state at the next moment is independent of the state at the past moment).

Inspired by Markov analysis, we use the time granularity of TKGs to discretely divide states. Then the basic model of RTFE can be expressed as:
\begin{equation}
    S_{t_{i+1}}=S_{t_{i}} \cdot P_{t_{i}}\label{Si_trans}
\end{equation}
where $S_{t_i}$ represents the state of $G_{t_i}$ and $P_{t_i}$ represents probability transition matrix to transform $S_{t_i}$ to that of $t_{i+1}$. A typical KG embedding learner uses its parameters $\theta$ and features $X$ to represent the semantic information of KG. Thus we approximate state vectors as:
\begin{equation}
    S_{t_{i}}:=\left[\theta_{t_i},X_{t_i}\right]\label{Si_defi}
\end{equation}

The idea of RTFE is to dynamically adjust $\theta$ and $X$ as the TKG changes while passing the information of each timestamp graph. We simply assume the features and parameters satisfy Markov Property:
\begin{equation}
    \begin{aligned}
        P(X_{t_{i+1}},\theta_{t_{i+1}}|X_{t_{1}},...,X_{t_{i}};\theta_{t_{1}},...,\theta_{t_{i}})&\\
        =P(X_{t_{i+1}},\theta_{t_{i+1}}|X_{t_{i}},\theta_{t_{i}})\label{markov_property}
    \end{aligned}
\end{equation}
where $X_{t_{i}}$ and $\theta_{t_{i}}$ denote features and parameters at time $t_{i}$. 

RTFE does not specify a model, but rather a training method for TKG completion. Existing SKGE methods and TKGE methods that follow the SKGE training pattern such as DE-SimplE (\citealp{goel2019diachronic}) and TComplEx (\citealp{lacroix2020tensor}) can potentially be utilized as the embedding component. The RTFE framework is illustrated in Figure \ref{fig:RTFE}. In section 3, we specify that RTFE how to use SKGE models for TKG completion. In section 4, we generalize RTFE to existing TKGE models to enhance their performance. 
\subsection{Preliminary training for static features}
Instead of training from scratch, RTFE uses SKGE as input to the first timestamp. In order to obtain the input features, the TKG is transformed into SKG $G_{static}$, which is obtained by merging the facts of each timestamp:
\begin{equation}
    \begin{aligned}
        G_{static}: & =\bigcup_{i=1}^{n}G_{t_i}\\
        &=\{(s,r,o)|(s,r,o,t_i)\in G_{t_i},G_{t_i}\in G\}\label{G_static}
    \end{aligned}
\end{equation}

Suppose the SKG embedding learner be $\theta$, which takes the knowledge graph $G$ (facts of $G$) and the feature $X$ (which can be predefined or randomly initialized) as inputs. Send $G_{static}$ and $X$ to $\theta$, and then get the updated feature $X$ after training, which will be the input to the first timestamp.

\subsection{Learning each timestamp recursively}
In TKGs, parameters $\theta$ and features $X$ should change with time (i.e., with the change of TKG). We find that, due to the continuity of facts, most of the facts are the same in the adjacent timestamps, while only a small number of facts changed. For discrete events, they influence the states of the surrounding entities, leading to the possibility that these entities may produce new facts. Therefore, model parameters and features fitting a certain timestamp provide a good starting point for the learning of the next timestamp.

Different from most neural network-based SKGE models (\citealp{schlichtkrull2018modeling}; \citealp{wu2019relation}) which only update $\theta$ during training, leaving the input features $X$ unchanged, we let $X$ be updated as well, to capture the temporal dynamics of entities and relations.

Therefore, in our framework RTFE, model parameters $\theta$ and input features $X$ are both updated in the way similar to equation (\ref{Si_trans}) during state transition:
\begin{equation}
    \left[\theta_{t_{i+1}},X_{t_{i+1}}\right]=\left[\theta_{t_i},X_{t_i}\right]\cdot P_{t_i}\label{theta_X_trans}
\end{equation}
where $\theta_{t_i}$ and $X_{t_i}$ denote the state vectors of $\theta$ and $X$ at time $t_i$ respectively; $P_{t_i}$ represent the probability transition matrix. To transform state vectors at $t_i$ to that at $t_{i+1}$, we approximate the state transition $P_{t_i}$ as the gradient update process of learning $G_{t_i}$ (i.e., updating according to the gradient of the loss function for several epochs):
\begin{equation}
    \theta_{t_{i+1}}=\theta_{t_i}-\alpha \cdot \nabla_{\theta}l(\theta_{t_i},X_{t_i},G_{t_i})\label{theta_update}
\end{equation}
\begin{equation}
    X_{t_{i+1}}=X_{t_i}-\alpha \cdot \nabla_{X}l(\theta_{t_i},X_{t_i},G_{t_i})\label{X_update}
\end{equation}
where $\alpha$ is the learning rate; $l$ is the loss function defined by the specified embedding learner; $\nabla_{\theta}$ is the gradient of $l$ with respect to $\theta$.

It must be pointed out that the state transition matrix in Markov analysis is fixed, so the above analysis method is generally applicable to short-term prediction. But the state vectors are different in different states and the gradient between the states is also different. Since the state vector is fixed in a specific state, a model can be established for each discrete state by the time interval of TKG. Then the gradient can be updated between states according to the difference of each state vector, to continue our framework.

RTFE recursively trains each timestamp according to equation (\ref{theta_update}) and (\ref{X_update}) and uses $\theta_{t_i}$ and $X_{t_i}$ to test $G_{t_i}^{'}$. Since RTFE is trained and tested by timestamp, only the latest parameters and features need to be stored, which shows good scalability for large TKGs. The framework RTFE is illustrated in Figure \ref{fig:RTFE} and the overall training and testing algorithm is shown in Algorithm 1.
\begin{figure}[t]
    \centering
    \includegraphics[scale=0.5]{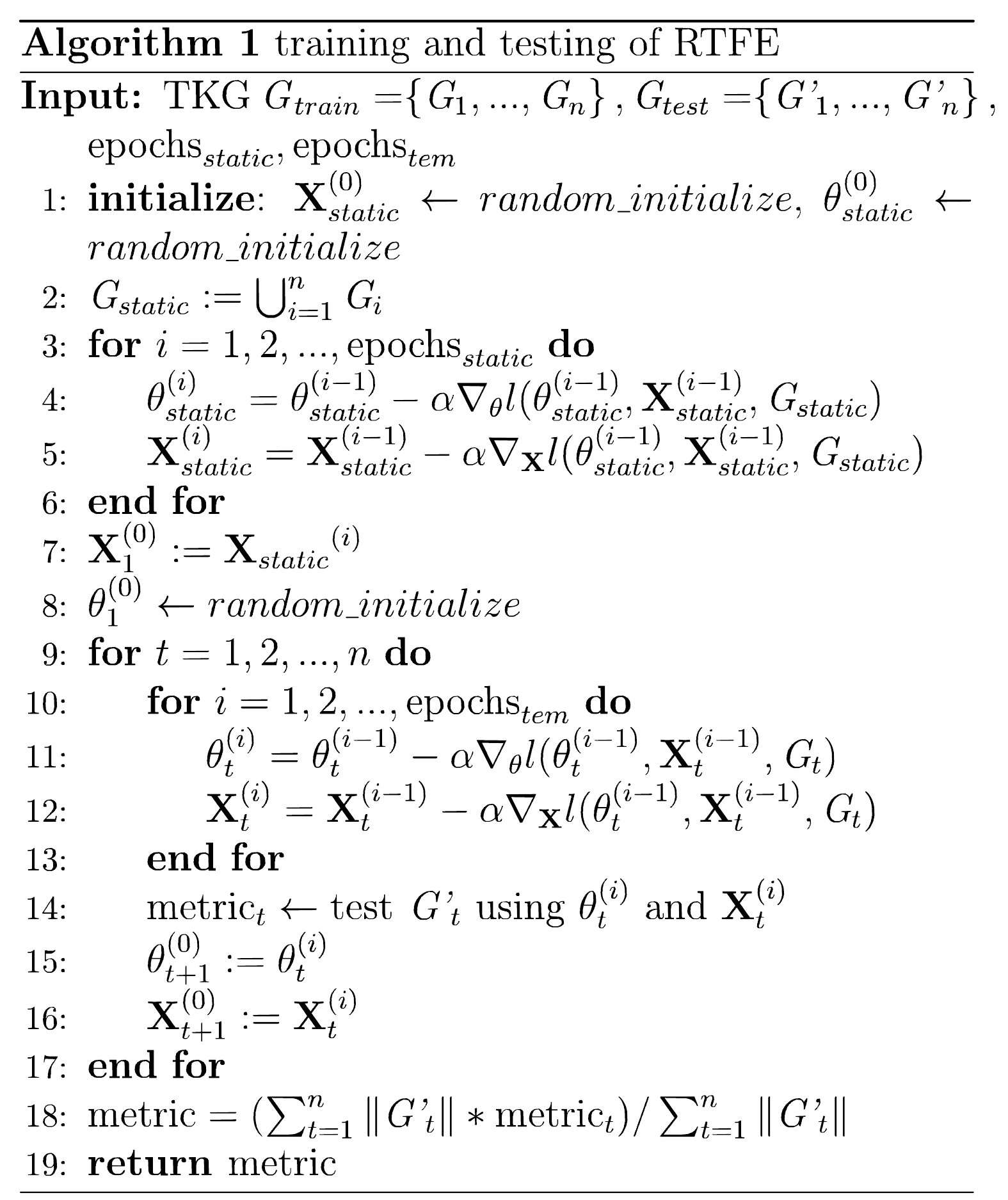}
    \label{RTFE_algorithm}
\end{figure}

\section{Transplanting SKGE/TKGE models}
\subsection{SKGE models}
For translation-based methods like TransE (\citealp{bordes2013translating}), TransD (\citealp{ji2015knowledge}), RotatE (\citealp{sun2019rotate}) and HAKE (\citealp{zhang2020learning}), they can be directly used as the embedding learner of RTFE without change of models.

For Graph neural network-based methods like RGCN (\citealp{schlichtkrull2018modeling}), we make its input feature do gradient update as well, so that the input features encode the information of each timestamp, so as to enhance the information transfer between timestamps. In addition, a residual connection is added to the network between the network inputs and outputs of each timestamp.

For RDGCN (\citealp{wu2019relation}) that was designed for entity alignment, in order to measure the plausibility of a triple \textit{(s, r, o)} for SKG completion, we design a distance function consisting of type distance and semantic distance:
\begin{equation}\label{RDGCN_LOSS}
    d(s,r,o)=d_{type}(s,r,o)+\lambda \cdot d_{seman}(s,r,o)
\end{equation}
\begin{equation}\label{RDGCN_TYPE}
     d_{type}(s,r,o)=|\left[\overline{X}_s^{E},\overline{X}_o^{E}\right]-\overline{X}_r^{R}|
\end{equation}
\begin{equation}\label{RDGCN_SEMANTIC}
\begin{aligned}
    d_{seman}(s,r,o)=|&(\left[\overline{X}_s^{E},\overline{X}_o^{E}\right]-\overline{X}_r^{R})[:d]\\-&(\left[\overline{X}_s^{E},\overline{X}_o^{E}\right]-\overline{X}_r^{R})[d:]|
\end{aligned}
\end{equation}
where $\overline{X}^{E} \in \mathbb{R}^{|V|\times d}$ and $\overline{X}^{R} \in \mathbb{R}^{|V|\times 2d}$ denotes output entity and relation representations.

\subsection{Enhancing TKGE models}
Since existing TKGE models such as DE-SimplE (\citealp{goel2019diachronic}) and TComplEx (\citealp{lacroix2020tensor}) for TKG completion follow the training pattern of SKGE models (i.e., think of TKG as a whole graph, not as a sequence of graphs), we can use them as the embedding leaner of RTFE. Specifically, we think their own training process as the preliminary training of RTFE. After TKGE models finish their own training process, we use the obtained features and parameters as the input to the learning of the first timestamp. Then RTFE trains the TKGE model recursively by equation (\ref{theta_update}) and equation (\ref{X_update}).

\subsection{Extensibility for future timestamps}
Since RTFE embeds each timestamp recursively, transforming form current state to the next state, it provides a way to complete upcoming future timestamps. Specifically, given a sequence of observed graphs of a TKG:  $G_{obs}=\{G_{t_1} ,…,G_{t_n}\}$ and a sequence of upcoming future graphs: $G_{fut}=\{G_{t_{n+1}},…,G_{t_{n+j}}\}$. We pre-train RTFE on $G_{obs}$, then embeds timestamp recursively to obtain the latest features  $X_{t_n}$ and parameters $\theta_{t_n}$. To complete $G_{t_{n+1}}$, we use equation (\ref{theta_update}) to equation (\ref{X_update}) to obtain $X_{t_{n+1}}$ and $\theta_{t_{n+1}}$ similarly. Then graphs of $G_{fut}$ can also be completed in this recursive way, without retraining of $G_{obs}$.

\section{Experiment}
\subsection{Experimental setup}
\textbf{Dataset:} We evaluated models on two fact datasets proposed by Hyte (\citealp{dasgupta2018hyte}): YAGO11k and Wikidata12k and three event datasets ICEWS14, ICEWS05-15 and GDELT (\citealp{goel2019diachronic}). The details of the five datasets are illustrated in appendix.\\
\textbf{Evaluation settings and metrics:} For entity prediction, we used mean reciprocal rank ($MRR$) and $Hits@1$, $Hits@3$, $Hits@10$ as metrics. Hits@n is defined as:
\begin{equation*}
    Hits@n=\frac{\sum_{fact \in test\_set}bool(rank(fact) \leq n)}{|test\_set|}
\end{equation*}

For relation prediction, we used mean rank ($MR$) $=\frac{\sum_{fact \in test\_set}rank(fact)\leq n}{\#test\_set}$ and $Hits@1$ as metrics since the number of relations is small. The rank of a test triple is obtained by replacing its head/tail/relation with remaining negative samples, and then evaluating the score rank of the original triple in all the replacement samples. Mean rank ($MR$) is the average rank of all test triples. And Mean reciprocal rank ($MRR$) is the average of the reciprocal ranks. 

For our RTFE framework, a timestamp-by-timestamp train-test mode was adopted. The total test result was a weighted average of all timestamp test results. For example, the final $MRR$ was calculated as:
\begin{equation*}
    MRR_{all}=\frac{\sum_{i}{|G_{t_i}|} \times MRR_{G_{t_i}}}{\sum_{i}{|G_{t_i}|}}
\end{equation*}
\textbf{Baselines:} We compared our framework RTFE to state-of-the-art TKGE models including t-TransE (\citealp{jiang2016towards}), Hyte (\citealp{dasgupta2018hyte}), DE-SimplE (\citealp{goel2019diachronic}), ATiSE (\citealp{xu2019temporal}), TComplEx (\citealp{lacroix2020tensor}) for TKG completion. Then we used several state-of-the-art SKGE models including TransE (\citealp{bordes2013translating}), TransD (\citealp{ji2015knowledge}), RGCN (\citealp{schlichtkrull2018modeling}), RDGCN (\citealp{wu2019relation}), RotatE (\citealp{sun2019rotate}) and HAKE (\citealp{zhang2020learning}) as the embedding learner of RTFE to perform TKG completion as well. And finally we use TKGE models as the embedding learner of RTFE to show the gain of performance.

\subsection{Entity prediction}
\begin{table*}
\centering
\resizebox{\textwidth}{!}{ 
\begin{tabular}{ccccccccc}
\hline
\textbf{Dataset} & \multicolumn{4}{c}{\textbf{Wikidata12k}} & \multicolumn{4}{c}{\textbf{YAGO11k}}\\
\hline
\multirow{2}{*}{Metric} & MRR & Hits@1 & Hits@3 & Hits@10 & MRR & Hits@1 & Hits@3 & Hits@10\\
 & tail head & tail head & tail head & tail head & tail head & tail head & tail head & tail head\\
\hline
t-TransE* & 17.2 & 9.6 & 18.4 & 32.9 & 10.8 & 2.0 & 15.0 & 25.1\\
Hyte & 21.4 14.3 & 12.0 8.0 & 23.6 14.5 & 41.8 26.1 & 14.7 6.0 & 2.9 0.2 & 19.9 9.1 & 35.0 13.4\\
ATiSE* & 28.0 & 17.5 & 31.7 & 48.1 & 17.0 & 11.0 & 17.1 & 28.8\\
TComplEx & 44.9 29.5 & 29.9 21.2 & 55.8 31.8 & 67.4 47.6 & 32.4 18.0 & 22.7 13.5 & 35.7 17.1 & 52.2 27.1\\
\hline
RTFE-TransE & 36.4 14.1 & 20.1 5.6 & 47.3 13.6 & 67.6 33.4 & 19.8 7.6 & 6.3 0.4 & 27.0 13.1 & 42.5 14.6\\
RTFE-TransD & 33.0 7.2 & 18.1 2.8 & 45.1 8.0 & 63.1 18.5 & 18.9 9.6 & 3.7 0.5 & 26.5 14.7 & 46.2 22.1\\
RTFE-RGCN & 27.6 15.6 & 20.8 8.6 & 30.8 17.4 & 41.2 31.9 & 20.3 14.8 & 16.4 12.5 & 21.2 14.5 & 28.1 19.0\\
RTFE-RDGCN & 43.5 17.0 & 36.2 13.2 & 46.7 18.6 & 58.1 23.9 & 20.9 10.0 & 8.8 2.6 & 27.1 13.3 & 43.0 21.5\\
RTFE-RotatE & 44.6 27.8 & 34.0 19.0 & 52.7 31.9 & 63.3 45.9 & 28.9 18.5 & 19.8 14.3 & 32.0 18.1 & 47.9 27.0\\
RTFE-HAKE & 35.7 27.3 & 24.7 19.4 & 43.1 30.0 & 54.6 42.7 & 30.8 \textbf{20.0} & 20.8 \textbf{14.6} & 34.1 \textbf{19.8} & 52.3 \textbf{31.3}\\
\hline
RTFE-TComplEx & \textbf{52.2 38.6}& \textbf{42.6 30.6}& \textbf{59.7 42.0} & \textbf{68.6 55.2}& \textbf{32.9} 18.7& \textbf{23.5} 14.1 & \textbf{35.9} 17.9 &\textbf{52.7} 28.2\\
\end{tabular}
}

\caption{\label{entity_predict_fact}
Entity prediction on continuous fact datasets: YAGO11k and Wikidata12k. Since the relations of these 2 datasets have typical “one-to-many” nature, the performance of tail prediction is better than that of head prediction.
}
\end{table*}

\begin{table*}
\centering
\resizebox{\textwidth}{!}{
\begin{tabular}{ccccccccccccc}
\hline
\textbf{Dataset} & \multicolumn{4}{c}{\textbf{ICEWS14}} & \multicolumn{4}{c}{\textbf{ICEWS05-15}}&\multicolumn{4}{c}{\textbf{GDELT}}\\
\hline
Metric & MRR & Hits@1 & Hits@3 & Hits@10 & MRR & Hits@1 & Hits@3 & Hits@10 & MRR & Hits@1 & Hits@3 & Hits@10\\
\hline
t-TransE* & 25.5 & 7.4 & -- & 60.1 & 27.1 & 8.4 & -- & 61.6 & 11.5 & 0.0 & 16.0 & 31.8\\
Hyte* & 29.7 & 10.8 & 41.6 & 65.5 & 31.6 & 11.6 & 44.5 & 68.1 & 11.8 & 0.0 & 16.5 & 32.6\\
DE-SimplE & 52.6 & 41.8 & 59.2 & 72.5 & 51.3 & 39.2 & 57.8 & 74.8 & 23.0 & 14.1 & 24.8 & 40.3\\
ATiSE* & 54.5 & 42.3 & 63.2 & 75.7 & 51.9 & 37.8 & 60.6 & 79.4 & -- & -- & -- & --\\
TComplEx & 54.7 & 44.5 & 60.8 & 73.2 & 58.3 & 48.0 & 65.0 & 78.7 & 21.7 & 12.8 & 23.1 & 37.2\\
\hline
RTFE-RotatE & 45.1 & 30.1 & 55.8 & 70.8 & 35.9 & 12.5 & 54.0 & 72.4 & 35.7 & 24.8 & 39.7 & 57.9\\
RTFE-HAKE & 50.3 & 40.2 & 55.7 & 70.0 &47.1 & 37.0 & 54.0 & 64.5 & \textbf{50.2} & \textbf{43.8} & \textbf{52.8} & \textbf{62.0}\\
\hline
RTFE-DE-SimplE & 57.3 & 49.4 & 62.1 & 71.7 & 58.7 & 50.6 & 63.7 & 73.0 & 42.9 & 35.9 & 45.5 & 55.9\\
RTFE-TComplEx & \textbf{59.2} & \textbf{50.3} & \textbf{64.6} & \textbf{75.8} & \textbf{64.5} &\textbf{55.3} &\textbf{70.6}& \textbf{81.1} & 29.7 & 21.2 & 31.9 & 46.4\\
\end{tabular}}
\caption{\label{entity_predict_event}
Entity prediction on discrete event datasets: ICEWS14, ICEWS05-15 and GDELT.
}
\end{table*}
Entity prediction is given a quadruple \textit{(s, r, o, t)}, to perform head entity prediction (i.e., to predict the plausibility of \textit{(?, r ,o, t)} ) , and performs tail entity prediction (i.e., to predict the plausibility of \textit{(s, r, ?, t)}). The plausibility of \textit{(s, r, o, t)} is ranked among all corrupted quadruples, while all true quadruples are excluded according to TransE's filtering protocol. The experimental results are shown in Table \ref{entity_predict_fact} and Table \ref{entity_predict_event} where results marked (*) are taken from reported results of Hyte and ATiSE.

Table \ref{entity_predict_fact} shows that on continuous fact datasets, both translation-based and graph neural network-based methods can be transplanted to RTFE, and the results are better than Hyte, indicating the generality and superiority of RTFE. For both TransE-based approaches, RTFE-TransE outperforms Hyte on all metrics (e.g., with the improvement of 15.0\% in tail $MRR$ on Wikidata12k) because RTFE takes advantage of the continuity of facts directly.

RotatE and HAKE are the most advanced translation-based approaches and RTFE-RotatE or RTFE-HAKE outperforms other methods, which demonstrates that our framework can preserve the excellent results of these methods over SKG. Besides, the performance of state-of-the-art TKGE model TComplEx is enhanced by RTFE, which shows the gain from our recursive training pattern.

Table \ref{entity_predict_event} shows that on discrete event datasets, RTFE also significantly improves the performance of TKGE models. Besides, on a dense dataset (i.e., with a small number of entities and a large number of facts) like GDELT, RTFE can take advantage of SKGE models such as HAKE.

\subsection{Relation prediction}
\begin{figure*}[t]
    \centering
    \includegraphics[width=1\textwidth]{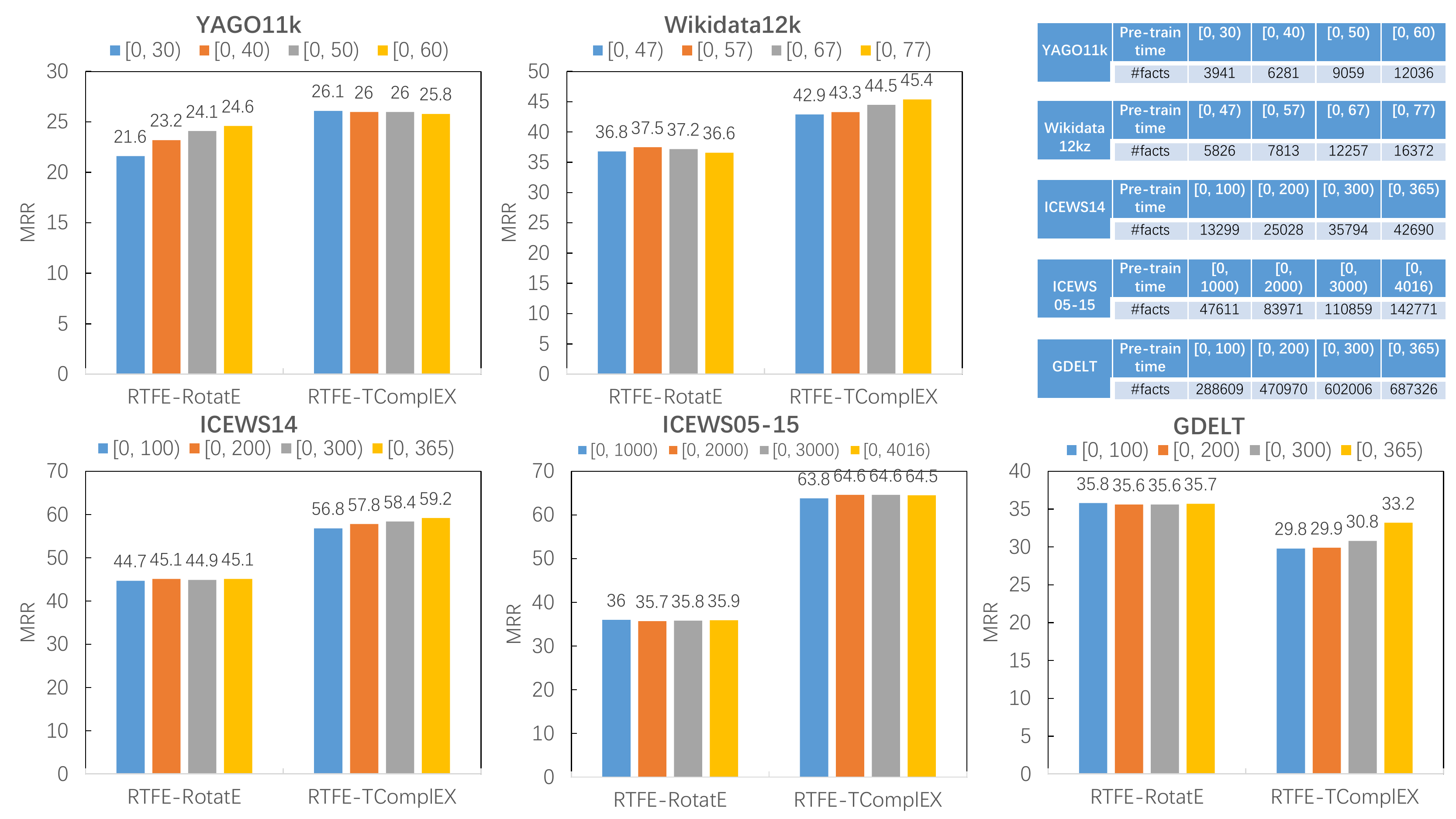}
    \caption{Extensibility validation experiment. Using equation (\ref{G_static}) to convert TKG to SKG, different time intervals of TKG are used for pre-training. Then all timestamps are trained and tested to validate the extensibility for timestamps unseen during pre-training. The horizontal axis represents the time interval for pre-training and the ordinate represents the MRR of test results for all timestamps.}
    \label{fig:extensibility}
\end{figure*}
Relation prediction is given a quadruple \textit{(s, r, o, t)}, to evaluate the plausibility of \textit{(s, ?, o, t)}. The experimental results are shown in Table \ref{tab:relation_predict}. RTFE-RDGCN$_{type}$ outperforms Hyte on YAGO11k that has only 10 relations (e.g., with the improvement of 12.5\% in $Hits@1$), which implies that type information plays an important role in this task. Since the number of relations between these two datasets is relatively small (10 and 24), the performance improvement is not obvious after adding semantic information (e.g., with the improvement of 1.3\% in $Hits@1$).

Hyte performed well on Wikdata12k. This may be attributed to its SKGE training pattern, which helped to capture applicable relation types between two entities from all the facts. In contrast, the timestamp of RTFE is trained by time, so only the facts of the current timestamp and information of last timestamp are directly utilized. To provide RTFE-RDGCN with more training data about relations, we added additional 30\% negative samples obtained by replacing relations of quadruples into the negative sample set: $\{(s,r^{'},o)|(s,r,o) \in G_t,(s,r^{'},o)\notin G\}$. We call this variant RTFE-RDGCN$_{rel}$, which improves the performance of relation prediction on Wikidata12k compared with RTFE-RDGCN (with the improvement of 9.1\% in Hits@1).

\subsection{Extensibility validation}
\begin{table}
\centering
\scalebox{0.90}{
\begin{tabular}{ccccc}
\hline
\textbf{Dataset} & \multicolumn{2}{c}{\textbf{YAGO11k}} & \multicolumn{2}{c}{\textbf{Wikidata12k}}\\
Metric & MR & Hits@1 & MR & Hits@1 \\ 
\hline
TransE* & 1.70 & 78.4 & 1.35 & 88.4\\
TransH* & 1.53 & 76.1 & 1.40 & 88.1\\
HoIE* &2.57 & 69.3 & 2.23 & 84.0\\
t-TransE* &1.66 & 75.5 & 1.97 & 74.2\\
Hyte* &1.23 & 81.2 & \textbf{1.13} & \textbf{92.6}\\
\hline
RTFE-TransE & 1.43 & 84.1 & 1.88 & 73.7\\
RTFE-RDGCN$_{type}$ & 1.19 & 93.7 & 1.77 & 82.9\\
RTFE-RDGCN & 1.11 & 95.0 & 1.36 & 83.2\\
RTFE-RDGCN$_{rel}$ &\textbf{1.10} & \textbf{96.4} & 1.20 & 92.3
\end{tabular}}
\caption{Relation prediction. For RTFE-RDGCNtype, $\lambda$ is set to 0, which only considered type distance. For RTFE-RDGCN, $\lambda$ is set to 0.2.}
\label{tab:relation_predict}
\end{table}
In order to verify the influence of pre-trained static features on RTFE’s entire TKG completion, we divide timestamps into four time intervals and perform pre-training of RTFE on them respectively. Then the pre-trained static features of these time intervals are used as inputs to RTFE to test the performance of entity prediction at all timestamps.

The experimental results are presented in Figure \ref{fig:extensibility}. Although a complete SKG is not provided for pre-training, RTFE still remains a similar performance, which verifies the framework's extensibility for future timestamps. So RTFE can be extended to future timestamps to some extent, without the re-training of former timestamps, which shows good lightness and immediacy.

\subsection{Ablation study}
\begin{table*}
\centering
\resizebox{\textwidth}{!}{
\begin{tabular}{ccccccccccc}
\hline
\textbf{Dataset} & \multicolumn{2}{c}{Wikidata12k} & \multicolumn{2}{c}{YAGO11k} & \multicolumn{2}{c}{ICEWS14} & \multicolumn{2}{c}{ICEWS05-15} & \multicolumn{2}{c}{GDELT}\\
Metric & MRR &Hits@1 & MRR & Hits@1 & MRR & Hits@1 & MRR & Hits@1 & MRR & Hits@1\\ 
\hline
RTFE-RotatE & \textbf{36.2} & \textbf{22.1} & \textbf{23.7} & \textbf{17.1} & \textbf{45.1} & \textbf{30.1} & \textbf{35.9} & 12.5 & \textbf{35.7} & \textbf{24.8}\\
\multicolumn{1}{r}{\textbf{w/o pre-train}} & 27.5 & 21.2 & 20.1 & 16.4 & 33.0 & 17.6 & 35.7 & \textbf{16.0} & 32.5 & 20.3\\
\hline
DE-SimplE & - & - & - & - & 52.6 & 41.8 & 51.3 & 39.2 & 23.0 & 14.1\\
RTFE-DE-SimplE & - & - & - & - & \textbf{57.3} & \textbf{49.4} & \textbf{58.7} & \textbf{50.6} & 42.9 & 35.9\\
\multicolumn{1}{r}{\textbf{w/o pre-train}} & - & - & - & - & 48.5 & 42.6 & 53.1 & 46.8 & \textbf{44.7} & \textbf{37.9}\\
\hline
TComplEx & 37.2 & 26.6 & 25.2 & 18.1 & 54.7 & 44.5 & 58.3 & 48.0 & 21.7 & 12.8\\
RTFE-TComplEx & \textbf{45.4} & \textbf{36.6} & 25.8 & 18.8 & \textbf{59.2} & \textbf{50.3} & \textbf{64.5} & \textbf{55.3} & 29.7 & 21.2\\
\multicolumn{1}{r}{\textbf{w/o pre-train}} & 41.9 & 34.4 & \textbf{26.5} & \textbf{20.0} & 54.4 & 41.6 & 63.2 & 54.1 & \textbf{34.2} & \textbf{25.9}\\
\hline

\end{tabular}}
\caption{Ablation study. w/o pre-train refers to RTFE without preliminary training for static futures.}
\label{tab:ablation}
\end{table*}
In this subsection, we explore the effects of preliminary training and recursive training. \textbf{w/o pre-train} refers to RTFE without preliminary training for static futures (i.e., only recursive training). As shown in Table \ref{tab:ablation}, RTFE generally outperforms \textbf{w/o pre-train}, which verfies the significance of preliminary training. For TKGE models, \textbf{w/o pre-train} is generally competitive with them. With preliminary training, RTFE gets a good starting point. So RTFE outperforms the original TKGE models. 

However, it's interesting to find that on GDELT \textbf{w/o pre-train}s outperform RTFE-DE-SimpleE and RTFE-TComplEx. This can be attributed to the special denseness of this dataset. RTFE can fit each timestamp well with its enough facts and recursive training. Pre-training provides a local optimum of all timestamps. In this case, it's not as good as the random initialization of the first timestamp 

\section{Related work}

There are two kinds of facts in the TKGs: continuous facts and discrete events. Continuous facts have temporal attributes \textit{(since… is true, until… ends)} like \textit{(Obama, President, United States, 2009-2016)}. And discrete events have temporal attributes \textit{(happens at…)} like \textit{(Obama, Inaugurated as, President, 2009)}.

In recent years, some work (\citealp{jiang2016towards}; \citealp{esteban2016predicting}; \citealp{tresp2017embedding}; \citealp{trivedi2017know};  \citealp{garcia2018learning}; \citealp{jain2020temporal}; \citealp{ma2019embedding}; \citealp{xu2019temporal}; \citealp{jin2019recurrent}; \citealp{liu2019context}; \citealp{wang2019hybrid}; \citealp{tang2020timespan}; \citealp{goel2019diachronic}; \citealp{xu2020tero}; \citealp{jain2020temporal}; \citealp{lacroix2020tensor};) began to use the time information to improve the KG completion or directly complete the TKG. Based on the fact or event they dealt with, we state representative TKGE methods as follows.

\textbf{(1) Event completion:} DE (\citealp{goel2019diachronic}) made the entity embedding into a function DEEMB that takes the time point as a variable. While DE transplanted SKG embedding methods to TKGs, it didn’t involve recent GNN-based SKG embedding methods. TComplEx (\citealp{lacroix2020tensor}) presented an extension of Complex (\citealp{trouillon2016complex}) by adding timestamp embedding into decomposition of tensors of order 4. ATiSE (\citealp{xu2019temporal}) incorporated time information into entity/relation representations by using Additive Time Series decomposition. 

\textbf{(2) Event prediction:} (\citealp{esteban2016predicting}) trained an event prediction model by using background information provided by KG and recent events. RE-NET (\citealp{jin2019recurrent}) models the event sequence as a temporal joint probability distribution. The method is trained on historical data and then, by sampling from the probability distribution, predicts the events of the future timestamp graph. GHNN (\citealp{han2020graph}) used Hawkes process to capture the dynamic of evolving graph sequences. Glean (\citealp{deng2020dynamic}) incorporated both relational and world contexts to capture historical information.

\textbf{(3) Continuous fact completion:} (\citealp{jiang2016towards}; \citealp{jiang2016encoding}) used the order of relations and temporal consistency constraints to improve completion but did not make the embedding space directly contain time information. (\citealp{garcia2018learning}) used RNN to learn the representation of temporal relations, but did not consider that the embedding of entities should also change over time. Hyte (\citealp{dasgupta2018hyte}) represented timestamps as hyperplanes, and projected the entities and relations onto these hyperplanes. Then, the facts of all timestamps are learned jointly using a translation-based score function.

\section{Conclusion}
We propose a framework RTFE for TKG completion. We have transplanted SKGE models to TKGs and enhance the performance of existing TKGE models. Experiments show that on five TKG datasets RTFE outperformed baselines and is extensible for future timestamps to some extent.

In the future, we will further deal with discrete events. Since events with adjacent timestamps are correlated, we plan to modify RTFE so that it can learn correlations (especially causality) of events. By modeling spatio-temporal dependency of TKG, events in future timestamps can be forecasted. Besides, we plan to deal with the task of predicting time validity of facts (\citealp{leblay2018deriving}).

\section*{Acknowledgements}
This work was supported in part by the National Science Foundation of China (Grant No. 61902034); Engineering Research Center of Information Networks, Ministry of Education.

\bibliography{anthology,custom}
\bibliographystyle{acl_natbib}

\clearpage
\appendix

\section{Details of datasets}
\label{sec:appendix_dataset}

\begin{table*}[t]
\centering
\resizebox{\textwidth}{!}{
\begin{tabular}{cccccccc}
\hline
\textbf{Type} & \textbf{Dataset} & \textbf{$\#$Entity} &\textbf{$\#$Relation} &\textbf{$\#$Timestamp} &\textbf{$\#$Train} &\textbf{$\#$Validation} &\textbf{$\#$Test}\\
\hline
\multirow{2}{*}{Fact} & YAGO11k & 10623 & 10 & 60 & 203858 & 21763 & 21159\\
& Wikida12k & 12554 & 24 & 77 & 239928 & 18633 & 17616\\
\hline
\multirow{2}{*}{Event} & ICEWS14 & 7128 & 230 & 365 & 72826 & 8941 & 8963\\
& ICEWS05-15 & 10488 & 251 & 4017 & 368962 & 46275 & 46092\\
& GDELT & 500 & 20 & 366 & 2735685 & 341961 & 341961\\
\hline
\end{tabular}
}
\caption{\label{dataset_details}
Statistics on YAGO11k, Wikidata12k, ICEWS14, ICEWS05-15, and GDELT.
}
\end{table*}

In YAGO11k and Wikidata12k, each fact was formatted as \textit{(s, r, o, start\_date, end\_date)}. Following the step of Hyte to divide the timestamps, we used a year with more than 300 occurrences as the timestamp boundary. So the timestamp could be formatted as (year, the next year with more than 300 occurrences). Then, the facts were divided into these timestamps to get the set of quadruples \textit{(s, r, o, t)}, where \textit{t} represents the timestamp of the fact (A original fact can appear in multiple adjacent timestamps after decomposition). The statistics of the five datasets are illustrated in Table \ref{dataset_details}.

\section{Examples of feature $X$ in RTFE}
Feature $X$ refers to the input vectors of entites/relations. E.g., for RGCN (\citealp{schlichtkrull2018modeling}), feature $X$ refers to the input hidden state $h^{(0)}$ at layer 0; For TransE (\citealp{bordes2013translating}), feature $X$ refers to the embeddings.

\section{Parameter settings}
\begin{table*}
\centering
\resizebox{\textwidth}{!}{
\begin{tabular}{cccccccc}
\hline
\textbf{Model} & \textbf{Entity dim} & \textbf{Relation dim} &\textbf{Batchsize} &\textbf{Neg ratio} &\textbf{Leanring rate} &\textbf{Pre-train epochs} &\textbf{Epochs per time}\\
\hline
TransE & 300 & 300 & $\#$triplets/100 & 1 & 0.01 & $\leq$ 1000 & $\leq$ 200\\
TransD & 50 & 50 & 200 & 1 & 0.0001 & 5000 & 200\\
RGCN & 500 & 500$\times$ 500 & 1400 & 10 & 0.001 &$\leq$ 6000 & $\approx$ 2000\\
RDGCN & 150 & 300 & Full batch & 10;50 & 0.001 & 1000 & 100;200\\
RotatE & 2000 & 1000 & 1024 & 256 & 0.0001 & 6000 & 300\\
\hline
DE-SimplE & 100 & 100 & 512 & 500 & 0.001 & 500 & 100\\
TComplEx & 256 & 256 & 1000 & 0 & 0.01 & 50 & 20\\
\end{tabular}
}
\caption{\label{parameter settings}
Parameter settings of embedding models.
}
\end{table*}

The same parameters are used for the SKGE method and its corresponding RTFE version. Their main parameters are shown in Table \ref{parameter settings}. Besides, for TransE, the number of batches is set to 100. For RGCN, dropout ratio is set to 0.2; the number of GCN layer is set 2 and a res-net layer is added between the two RGCN layers. For RDGCN, $\lambda$ is set to 0.2.\\

\end{document}